\documentclass[sigconf, anonymous=false, authorversion=true]{acmart}

\usepackage{booktabs} 

\usepackage{subfig}

\usepackage[utf8]{inputenc}

\copyrightyear{2018} 
\acmYear{2018} 
\setcopyright{acmcopyright}
\acmConference[HAI '18]{6th International Conference on Human-Agent Interaction}{December 15--18, 2018}{Southampton, United Kingdom}
\acmBooktitle{6th International Conference on Human-Agent Interaction (HAI '18), December 15--18, 2018, Southampton, United Kingdom}
\acmPrice{15.00}
\acmDOI{10.1145/3284432.3284438}
\acmISBN{978-1-4503-5953-5/18/12}

\begin{document}
\title{
Bayesian Inference of Self-intention Attributed by Observer
	}
\renewcommand{\shorttitle}{
Bayesian Inference of Self-intention Attributed by Observer
}
\author{Yosuke Fukuchi}
\affiliation{Keio University, Yokohama, Japan}
\email{fukuchi@ailab.ics.keio.ac.jp}

\author{Masahiko Osawa}
\affiliation{Keio University, Yokohama, Japan}
\affiliation{Research Fellow of Japan Society for the Promotion of Science, Tokyo, Japan}
\email{mosawa@ailab.ics.keio.ac.jp}

\author{Hiroshi Yamakawa}
\affiliation{Dwango Artificial Intelligence Laboratory, Tokyo, Japan}
\affiliation{The Whole Brain Architecture Initiative, Tokyo, Japan}
\email{hiroshi\_yamakawa@dwango.co.jp}

\author{Tatsuji Takahashi}
\affiliation{Tokyo Denki University, Tokyo, Japan}
\affiliation{Dwango Artificial Intelligence Laboratory, Tokyo, Japan}
\email{tatsujit@mail.dendai.ac.jp}

\author{Michita Imai}
\affiliation{Keio University, Yokohama, Japan}
\email{michita@ailab.ics.keio.ac.jp}
\renewcommand{\shortauthors}{Y. Fukuchi et al.}

\begin{abstract}
	Most of agents that learn policy for tasks with reinforcement learning (RL) lack the ability
	to communicate with people, which makes human-agent collaboration challenging.
	We believe that, in order for RL agents to comprehend utterances from human colleagues,
	RL agents must infer the mental states that people attribute to them
	because people sometimes infer an interlocutor's mental states and communicate on the basis of this mental inference.
	This paper proposes PublicSelf model, which is 
	a model of a person who infers how the person's own behavior appears to their colleagues.
	We implemented the PublicSelf model for an RL agent in a simulated environment and examined
	the inference of the model by comparing it with people's judgment.
	The results showed that the agent's intention that people attributed to the agent's movement was correctly inferred 
	by the model in scenes
	where people could find certain intentionality from the agent's behavior.
\end{abstract}

%
%



\maketitle

\section{Introduction} \label{intro}
The development of reinforcement learning (RL) makes it possible to create agents 
that are capable of learning how to act instead of being programmed by designers.
Because RL agents can obtain policy by learning, it is possible to realize adaptive agents that work on tasks by trial and error.
In particular, with the introduction of deep learning, RL is expected to be applied to agents such as industrial and domestic robots
that act in the complex real world.

In general, however, 
collaboration between people and RL agents has many problems.
Natural language is one of the major media allowing people to interact with others. 
Thus, developing a mechanism for RL agents to comprehend people's utterances is an important step toward realizing human-friendly RL agents.

Many previous studies have focused on autonomous agents that comprehend people's utterances~\cite{nl_comprehension2, ishikawa-rinko}.
While people often use expressions that cannot be understood out of context, 
Sonja~\cite{Sonja} is an agent that can comprehend such context-dependent expressions.

\begin{figure}[tbp]
  \begin{center}
  	\includegraphics[clip,width=0.98\linewidth]{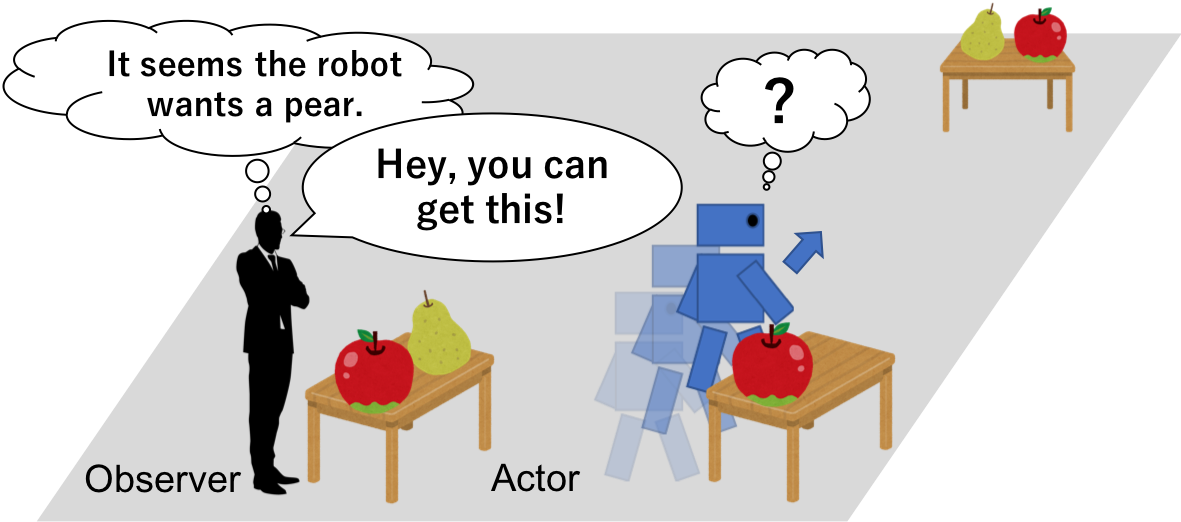}
  	\caption{Example of the scenario considered in this paper.
	  The utterance of the person to the robot is based on the mental states attributed to the robot by the person.
	  The robot needs to estimate the attributed mental states in order to comprehend the person's utterance.}
  	\label{fig:1}
  \end{center}
\end{figure}

Figure \ref{fig:1} shows an example scenario of the situation considered in this paper.
There are a robot and its colleague.
The colleague has an apple and a pear, while the robot has only an apple.
The robot moves toward the back side, and the colleague says to the robot, "Hey, you can get this."
If we were the robot, we would think the colleague was talking about the pear he had, and not about the apple.
We can interpret the utterance because we can infer that the colleague thinks the robot has a certain mental state 
even if the robot does not actually have such mental states as what people have,
and therefore assumes that "The robot {\it intends} to get a pear" and "The robot does not {\it want} an apple anymore."
Therefore, in order to comprehend the person's utterance, the robot also has to infer the person's inference about the robot.
In other words, the robot needs to infer how its behavior appears to the others.
Sonja's system assumes that the knowledge of Sonja's action target is shared with the colleague, 
and does not consider the mental inference by the colleague.

In this paper, we propose PublicSelf model.
PublicSelf model is a model of a person that infers another's inference of the person's mental state.
We also implemented the PublicSelf model for an RL agent in a simulated environment.
Implementation experiments showed that the model could estimate a human observer's inference of the person's intention
in scenes where people could discern a certain intentionality from the agent's behavior.

This paper is structured as follows.
Section 2 describes and formalizes the settings of the problem, and presents the background of the PublicSelf model.
Section 3 proposes the PublicSelf model.
Section 4 describes our implementation of the PublicSelf model for an agent in a simulated environment, and shows some examples of the model's inferences.
Section 5 describes an experiment to compare the inference of the PublicSelf model with a human observer's judgment.
Finally, section 6 concludes this paper.

\newcommand{\BEL}{{\rm BEL}}
\newcommand{\INTEND}{{\rm INTEND}}
\newcommand{\DONE}{{\rm DONE}}
\newcommand{\get}{{\rm get}}
\newcommand{\apple}{{\rm apple}}
\newcommand{\pear}{{\rm pear}}

\section{Background} 
\subsection{Mind-reading in human conversation}
\label{2.1}

Theory of mind is the ability to {\it read minds}, attributing a mental state such as a belief, a desire, a plan, or an intention to someone 
so as to understand and predict their behavior~\cite{tom}.
Mind-reading is considered to be one of the important elements for achieving social interaction.

We sometimes mind-read an interlocutor and communicate on the premise that the interlocutor has the mental state we infer. 
For example, \cite{ono} showed that people could understand the content of a robot's unclear utterance by considering 
the intention that they attributed to the robot.
In the example given in section 1, the person watches the behavior of the robot and attributes its intention to get a pear.
Thus, the person says "You can get this" based on the attributed mental state. 
We, on the other hand, can comprehend the person's utterance by inferring the intention the person attributes to the robot and 
assuming that the person intends to help the robot.
As seen here, people sometimes regard the results of mind-reading as the context when talking and interpreting someone's utterance
whether or not the interlocutor actually has mental states like people.

We formalize the process of people's thinking in the example scenario given in section 1 using Belief-Desire-Intention (BDI) logic~\cite{cohen}.
Here, we call the person an observer, and the robot an actor.
The utterance of the observer is based on the idea that the actor has intention $\iota^1$.

\begin{equation}
	(\BEL \; observer \; (\INTEND \; actor \; \iota^1)),
\end{equation}
where $(\BEL \; i \; \phi)$ means agent $i$ believes $\phi$, and $(\INTEND \; i, \; \phi)$ means agent $i$ intends to achieve $\phi$. 
The superscript $1$ of $\iota^1$ indicates that the variable is an inference of another person's mental state. 

The observer infers intention $\iota^1$ from the set of possible intentions for actor.
The observer chooses $\iota^1$ based on observations $o_{:t}$ and the actor's actions $a_{:t}$ at time $0, 1,..., t$.
In the example of section 1, the observer can be considered to attribute intention $\iota_p$, the intention to get a pear.
\begin{equation}
	\iota_a = (\get \; actor \; \apple).
\end{equation}
\begin{equation}
	\iota_p = (\get \; actor \; \pear).
\end{equation}

In order for the actor to comprehend the observer's utterance, the actor has to infer the intention attributed to them by the observer.
The inference is based on the actor's own observations and actions.
Let $\iota^2$ be the intention the actor infers that the observer attributes to them.
The superscript $2$ means that the variable is an inference of the self's mental state attributed by others.

\begin{equation}
(\BEL \; actor \; (\BEL \; observer \; (\INTEND \; actor \; \iota^2))).
\end{equation}

Moreover, the actor assumes that the observer's utterance was based on the observer's intention to help the actor.
\begin{equation} \label{trust}
(\BEL \; actor \; (\INTEND \; observer \; (\DONE \; actor \; \iota^2))),
\end{equation}
where $(\DONE \; i \; \phi)$ means that agent $i$ has just achieved $\phi$.

We can consider formula \ref{trust} to represent that the actor trusts the observer.
If the actor assumes the observer intends to obstruct the actor, formula \ref{trust} is changed to formula \ref{distrust}.
\begin{equation} \label{distrust}
(\BEL \; actor \; (\INTEND \; observer \; (\DONE \; actor \; \lnot \iota^2))).
\end{equation}

In either case, the actor can interpret the observer's utterance by choosing the interpretation that supports formula \ref{trust} or \ref{distrust}.
Conversely, the actor needs to be able to infer $\iota^2$ in order to interpret utterances based on the speaker's mind-reading.

\subsection{Adaptive action selection of reinforcement learning agents by considering people's utterance}
\label{2.2}
Our final goal is to propose a mechanism of RL agents that can adaptively select actions by considering people's utterances.
Adaptive action selection by RL agents involves not only comprehending context-dependent expressions from people, 
but also considering whether or not the agents should change their policy on the basis of people's utterances. 

Understanding the internal states of an RL agent, such as their goals, plans, beliefs, desires, and intentions, 
is a challenging problem because the agent does not have explicit representations for them.
This problem becomes a more important domain of research with the introduction of deep learning~\cite{IBE}.
The incomprehensibility of RL agents also raises the problem of the RL agent determining how to act while taking people's utterances into account. 

Sonja~\cite{Sonja} is an artificial agent which 
can comprehend an context-dependent expression such as "No, the other one."
If Sonja's observer says "No, the other one" when Sonja moves to an amulet, Sonja will begin searching for another amulet, 
or when Sonja moves toward a ladder, Sonja will change its direction to another ladder. 

When Sonja receives the message "No, the other one," 
its control logic handles the message by changing Sonja's target to another one.
Sonja's approach is based on two assumptions:
(i)~The observer knows what the actor is going to do, and
(ii)~The actor knows what the actor is going to do.

Suppose an actor is choosing actions based on intention $\iota^*$.
\begin{equation}
	(\INTEND \; actor \; \iota^*).
\end{equation}
The superscript $*$ indicates that the variable is an agent's actual internal state.

%
An observer speaks to an actor based on $\iota^1$, the inference of the actor's intention. 
If the actor's decision making process is explicit, 
it is reasonable to consider that the observer's inference matches the actor's true intention ($\iota^1 = \iota^*$)
because it is not very difficult for the observer to access the actor's actual internal states.
However, 
when it comes to an RL agent, it is difficult for a human observer to access the agent's internal states embedded in the agent's policy.
We do not even know whether the policy of an RL agent has what we call intention.
What the observer can do is only to infer the actor's intention from knowledge, context, and observations of the actor's behavior.
Thus, the observer may speak based on a false belief that the agent has intention $\iota^1 (\neq \iota^*)$. 
If the observer is misinterpreting the actor's intention, the actor may as well not take the observer at their word.

In addition, an RL agent also does not know $\iota^*$ explicitly. 
Therefore, the actor cannot judge whether the observer's belief on the actor is correct.

In order to realize RL agents that can adaptively select actions by considering people's utterances,
an RL agent should be able to infer two intentions, $\iota^2$ and $\iota^0$. 
$\iota^0$ is the introspective inference of the agent's intention by the agent.
\begin{equation}
	(\BEL \; actor \; (\INTEND \; actor \; \iota^0)).
\end{equation}
The superscript $0$ indicates that the variable is a self inference of their own mental states. 

$\iota^2$ helps the agent interpret the observer's utterance, 
and a comparison between $\iota^2$ and $\iota^0$ enables the agent to judge whether what the observer says has value.

\subsection{Self-awareness}
\label{2.3}
In the field of psychology, self-awareness is the ability of people to recognize the self as an object of attention~\cite{obj_sa}.

Self-awareness is considered to have two aspects: public self-awareness and private self-awareness~\cite{pub_priv_sa2, pub_priv_sa}.
Private self-awareness is a personal belief in the self acquired by the introspection of self thoughts and feelings.
Public self-awareness, on the other hand, is a belief in the self as a social object, and involves how the self appears to others.

The idea of self-awareness can be associated with the self inferences of an actor's intention:
\begin{itemize}
	\item Private self-awareness is the inference of the self's actual mental states, and involves the inference of $\iota^0$.
	\item Public self-awareness is the inference of the self's mental states attributed by an observer, and involves the inference of $\iota^2$.
\end{itemize}

We believe that RL agents need to be {\it self-aware} in these two ways to determine their actions while considering people's utterances.
The PublicSelf model can be considered a model of public self-awareness for RL agents. 

\subsection{Bayesian modeling of theory of mind}
Previous studies have modeled the computational mechanisms of 
the human mind-reading ability as a Bayesian inference (Fig. \ref{fig:2} left)~\cite{baker, tom2, btom}.
In this paper, we collectively call these approaches the Bayesian Theory of Mind (BToM).
One of the targets in the BToM research field is modeling a human observer who 
attributes mental states to an actor while watching the actor's behavior. 
In a typical problem setting, an observer can observe the whole environment, including the actor in the environment, 
and attributes mental states such as the actor's belief $b_t^1$, desire $r_t^1$, and intention $\iota_t^1$ based on the environment states
$s_{:t}$ and the actor's actions $a_{:t}$ at time $0,1,...,t$.
Each variable of the mental states can be estimated as a probability.
\begin{equation}
	\label{btom}
	\begin{split}
		&P(b_t^{1}, r^{1}_t, \iota_t^{1}|s_{:t}, a_{:t}) \\
		& \propto \Sigma_{o^{1}_t, b^{1}_{t-1}, r^1_{t-1}, \iota^1_{t-1}} P(o_t^{1}|s_t) 
		\cdot P(b_t^{1} | o_t^{1}, b_{t-1}^{1}) \cdot P(a_t | b^1_t, \iota_t^{1}) \\
		& \; \; \cdot P(\iota_t^{1} | b_t^{1}, r^{1}_{t-1}, \iota_{t-1}^{1}) \cdot P(s_t | s_{t-1}, a_{t-1}) \\ 
		& \; \; \cdot P(b_{t-1}^{1}, r^{1}_{t-1}, \iota_{t-1}^{1} | s_{:t-1}, a_{:t-1}),
	\end{split}
\end{equation}
where $o^1$ is an observation that the observer infers the actor observes at time $t$.
The probability of each variable can be calculated using a forward algorithm~\cite{forward}.
The PublicSelf model is based on the BToM concept.

We can consider the BToM to be a model of an observer who monitors an actor's behavior
and the proposed PublicSelf model to be a model of an actor that infers an observer's inference of the actor's intention.
While the inference of the BToM is based on a third-person's perspective of an observer,
the PublicSelf model infers an actor's mental states attributed by the observer from the actor's own perspective.
The intention which the BToM model infers is $\iota^1$, and that of the PublicSelf model is $\iota^2$.

\section{Bayesian Public Self-awareness Model}
This paper proposes PublicSelf model.
The PublicSelf is a Bayesian inference model of public self-awareness
and estimates the probability of an actor's mental states attributed by an observer. 

\begin{figure}[tbp]
  \begin{center}
  	\includegraphics[clip,width=0.9\linewidth]{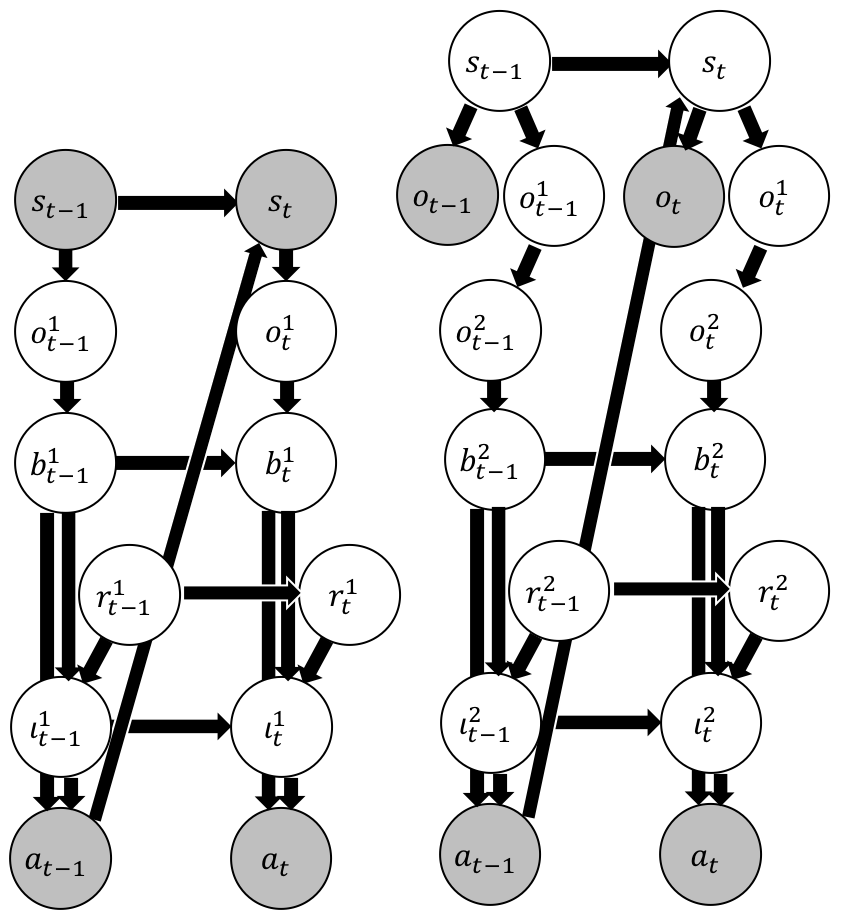}
	  \caption{Difference between Bayesian networks for typical BToM model (left) and PublicSelf model (right).
	  The subject of the BToM model is an observer, and the object is an actor.
	  In the case of the PublicSelf model, the relationship between the subject and the object is reversed.
	  The gray colored circle indicates that the variable is observable to the subject, while white means it is hidden.
	  The superscript numbers of each variable have the following meanings:
	  1 means that the subject attributes the variable to the object, and
	  2 means that the subject estimates the variable attributed to the subject by the object.}
  	\label{fig:2}
  \end{center}
\end{figure}

Figure \ref{fig:2} shows the Bayesian networks of the BToM model and the PublicSelf model.
In the BToM, observable variables are states of the environment $s$ and actions of the actor $a$.
However, because the PublicSelf model focuses on the inference by the actor, the observable variables are only what the actor can observe, that is, 
the observations of the actor $o$ and actions $a$.
The observer and actor observe the environment based on their own limited observation spaces.

The probabilities of the actor's attributed belief $b_t^2$, desire $r^2$, and intention $\iota^2_t$ can be estimated in a manner similar to that for the BToM:

\begin{equation}
	\label{eq:bpsa}
	\begin{split}
		& P(b_t^{2}, r^{2}_t, \iota_t^{2}|o_{:t}, a_{:t}) \\
		& \propto \Sigma_{s_t, o^1_t, o^{2}_t, b^{2}_{t-1}, r^2_{t-1}, \iota^2_{t-1}} P(o_t | s_t) \cdot P(o_t^{1} | s_t) \cdot P(o_t^{2} | o_t^{1}) \cdot P(b_t^{2} | o_t^{2}, b_{t-1}^{2}) \\
		& \; \; \cdot P(a_t | b^2_t, \iota_t^{2}) \cdot P(\iota_t^{2} | b_t^{2}, r^{2}_{t-1}, \iota_{t-1}^{2}) \cdot P(s_t | s_{t-1}, a_{t-1}) \\
		& \; \; \cdot P(b_{t-1}^{2}, r^{2}_{t-1}, \iota_{t-1}^{2} | o_{:t-1}, a_{:t-1}).
	\end{split}
\end{equation}

The crucial point of the PublicSelf model is to consider the asymmetry of the information between an actor and an observer.
For example, even if an actor observes an apple and moves toward it, the observer cannot read the actor's intention
unless the observer can observe both the actor and the apple.
The PublicSelf model emulates how the behavior of the actor appears to the observer in order to infer the mental states attributed by the observer.

\section{Implementation for RL agent in simulated environment}
\subsection{Simulated environment}
\label{implementation}
\begin{figure}[tbp]
	\begin{center}
		\subfloat[\label{fig:whole}]{
			\includegraphics[clip,width=.98\linewidth]{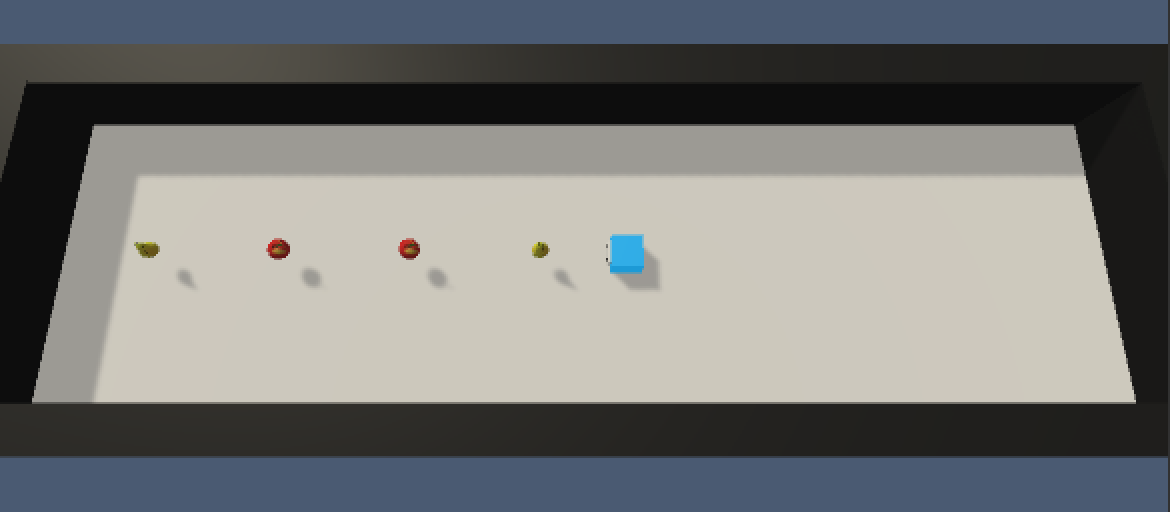}
		}
		\vspace{0.01cm}
		\subfloat[\label{fig:agent}]{
			\includegraphics[clip,width=.27\linewidth]{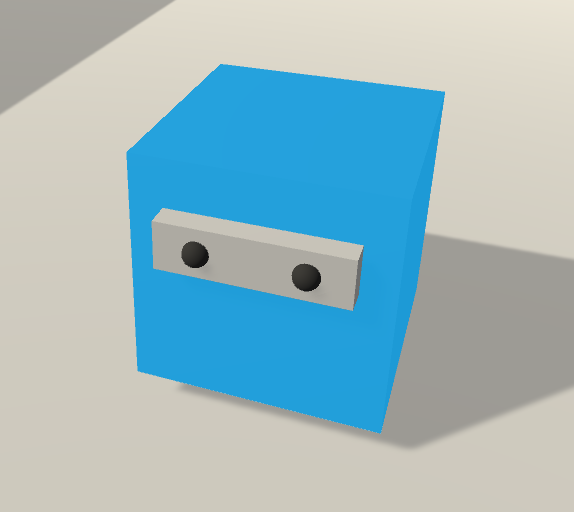}
		}
		\subfloat[\label{fig:fp}]{
			\includegraphics[clip,width=.31\linewidth]{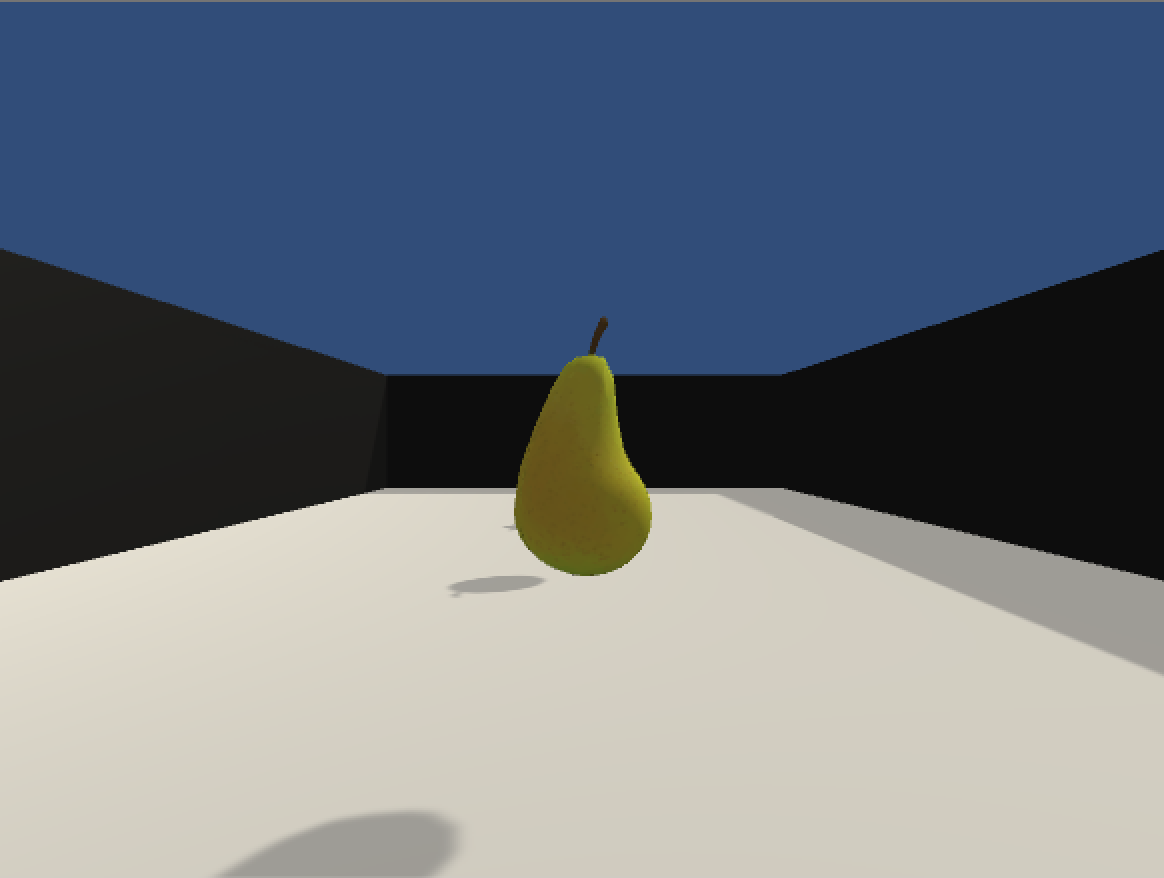}
		}
		\subfloat[\label{fig:hp}]{
			\includegraphics[clip,width=.37\linewidth]{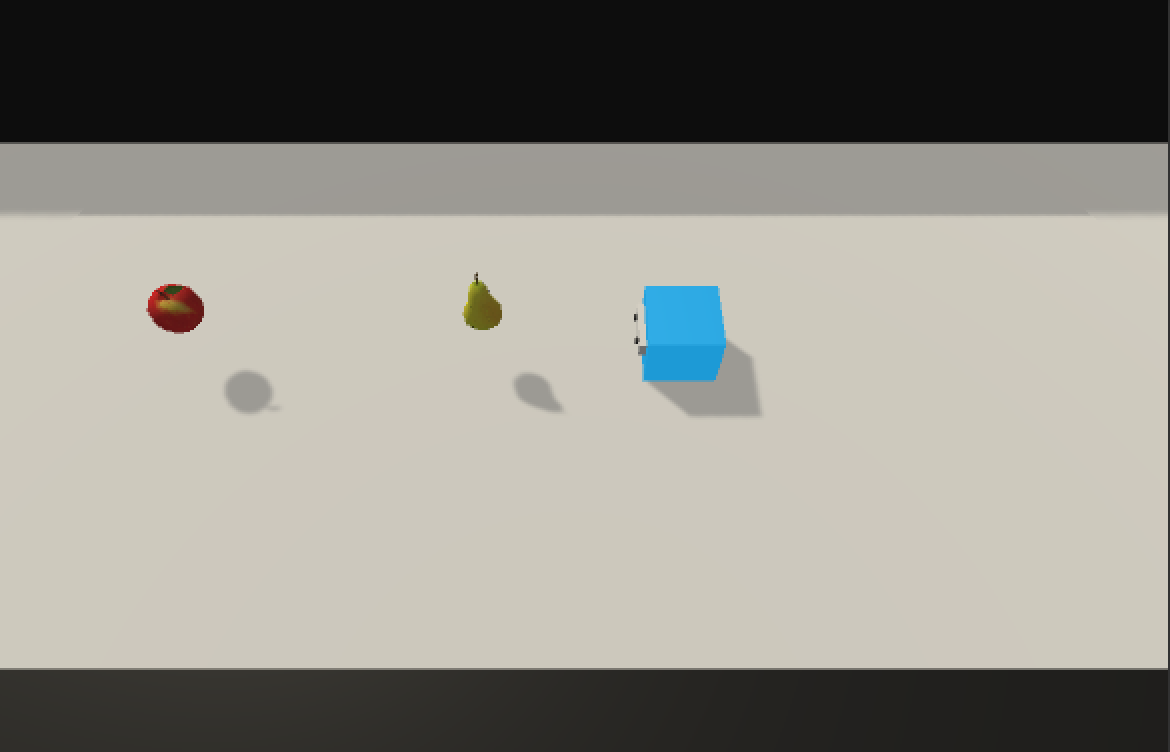}
		}
		\caption{ Simulated environment.
		(a) Bird's eye view of entire field. (b) Visual appearance of the agent (actor).
		(c) First-person view of the actor. (d) First-person view of the observer. 
		The observable area of the observer is limited to the center of the field.
		}
		\label{fig:3}
	\end{center}
\end{figure}

This paper introduces an implementation of the PublicSelf model for an RL agent (Fig. \ref{fig:agent}) 
in a simulated environment (Fig. \ref{fig:whole}).
In the environment considered, there are always five objects: two apples, two pears, and the agent (actor).
The objects spawn at random locations 
at the beginning of each episode.
The actor selects its action based on observation from a first-person view camera (Fig. \ref{fig:fp}).
The action space of the actor consists of five actions: moving forward, moving backward, turning clockwise, turning counterclockwise, and doing nothing. 

\subsection{Implementation}
In the current implementation, we did not calculate all of the probabilities in equation \ref{eq:bpsa}; some were updated deterministically.
We assumed that the observer observed the environment from a predefined fixed viewpoint (Fig. \ref{fig:hp}), and that the actor knew the area of the observer's view.
The view of the actor assumed by the observer was also predefined by the angle of view and distance from the front of the actor.

Environmental state $s$ consists of the velocity and direction of the actor and the locations of the objects on the field.
Belief $b(s)$ is the probability that the environment state is $s$ under the actor's observable variables ($o$ and $a$).

\begin{equation}
	b(s) = P(s | o_{:t}, a_{:t})
\end{equation}

We assume that both the actor and observer know there are always two apples and two pears on the field,
and consider that there are two possible desires attributed by the observer to the actor: $r_a$ and $r_p$.
\begin{eqnarray}
	r_a =\left\{
		\begin{array}{ll}
		1.5 & \text{if the actor touches an apple} \\
		-1.5 & \text{if the actor touches a pear} \\
		-0.002 & \text{else} \\
		\end{array} \right.
\end{eqnarray}
$r_p$ is constituted by replacing the apple and pear of $r_a$. 
We also assume that there are two possible intention: $\iota_a$ and $\iota_p$.
We assume that desires $r_a$ and $r_p$ respectively generate intentions $\iota_a$ and $\iota_p$, that is, 
$P(\iota_a | r_a) = P(\iota_p | r_p) = 1$.

The states of the simulation environment $s$ are continuous, so there are infinite possible worlds.
For the calculation, we reduce the possible states by considering the environment to be a grid world with seven rows and twenty-five columns.
The number of possible combinations where the fruits locate can be located is $\begin{pmatrix} 175\\ 2\end{pmatrix} \times \begin{pmatrix} 173\\ 2\end{pmatrix}  = 226,517,550$. 

Belief $b^2$ is initialized as the uniform probability distribution of all the possible states, 
and updated by rejecting states that contradict the observations overlapped between the actor and the observer. 
This means that the observer updates the belief attributed to the actor only if the observer can see what the actor observes.
If the actor passes by an apple to the right, disappears from the observer's view for a while, and returns,
people can infer that there are no pears in the right unobservable field, but this deterministic implementation cannot handle this kind of inference.

Equation \ref{eq:bpsa} has an element $P(a | b^2_t, \iota_t^2)$, which is the probability of selecting action $a$ under an actor's intention $\iota^2_t$ and belief $b^2_t$.
We need to calculate the possibilities under various possible intentions and beliefs,
but the actor's own policy, which handles first-person view observations as input, makes it hard to emulate various possibilities. 
Therefore, we prepared another policy $\pi_{meta}$, whose input is not visual images but numerical vectors consisting of the actor's velocity 
and the relative position between the actor and the fruits
using A3C, which is a deep reinforcement learning method~\cite{a3c}.
$\pi_{meta}$ calculates probability of choosing each action under certain belief, reward, and intention $P(a|b,\iota)$. 

\subsection{Intention estimation by implemented PublicSelf model}\label{3.1}

We verified the inference of an RL agent's intention using the PublicSelf model.
The actor has two policies, $\pi_a$, which was learned to move to apples with A3C, and $\pi_p$ to move to pears. 

\begin{figure}[tbp]
	\begin{center}
		\subfloat[$t=1$ (observer's perspective)\label{fig:result1_1}]{
			\includegraphics[clip,width=.49\linewidth]{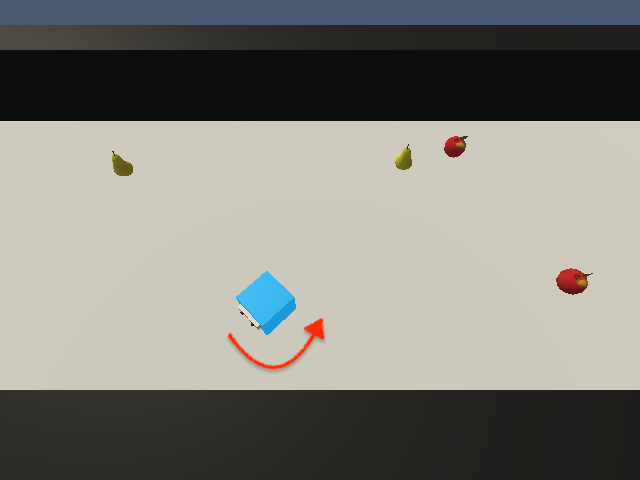}
		}
		\subfloat[$t=13$\label{fig:result1_2}]{
			\includegraphics[clip,width=.49\linewidth]{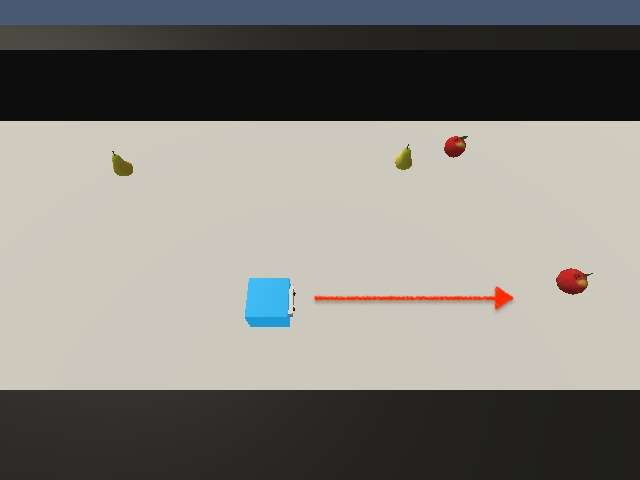}
		}
		\vspace{0.01cm}
		\subfloat[$t=18$\label{fig:result1_3}]{
			\includegraphics[clip,width=.49\linewidth]{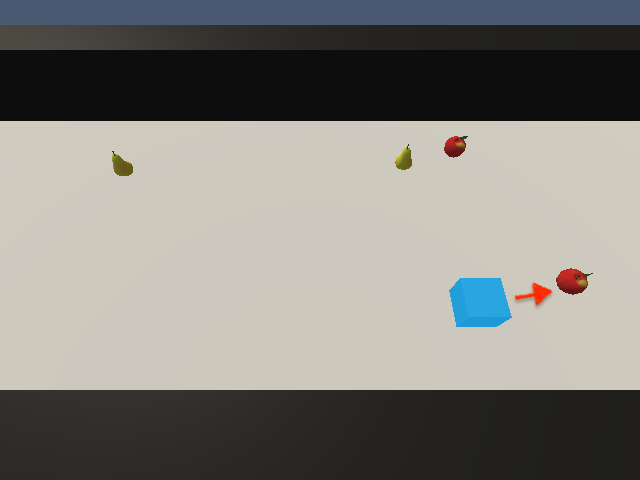}
		}
		\subfloat[probability \label{fig:result1_4}]{
			\includegraphics[clip,width=.49\linewidth]{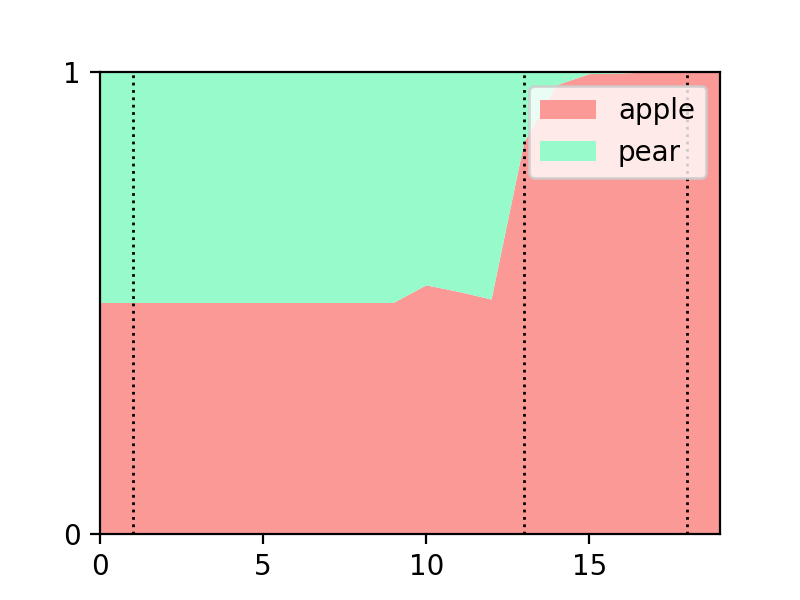}
		}
		\caption{
			The actor's behavior and the probability of the actor's intention $\iota^2$ were estimated by the PublicSelf model in episode 1.
			Here, the actor showed movement that made it relatively simple for an observer to infer the actor's intention.
			(a) First, the actor turned counterclockwise.
			(b) Then, the actor saw an apple in sight and began to move in a straight line.
			(c) Finally, the actor reached the apple after adjusting its direction of movement.
			(d) The probability of the actor's intention $\iota^2$ was estimated by the PublicSelf model.
			The model increased the probability as soon as the actor moved toward the apple.			
		}
		\label{fig:result1}
	\end{center}
\end{figure}

\begin{figure}[tbp]
	\begin{center}
		\subfloat[$t=1$\label{fig:result3_1}]{
			\includegraphics[clip,width=.49\linewidth]{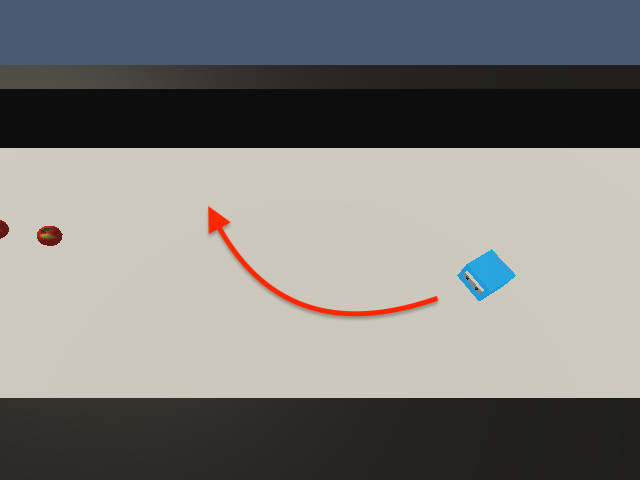}
		}
		\subfloat[$t=32$\label{fig:result3_2}]{
			\includegraphics[clip,width=.49\linewidth]{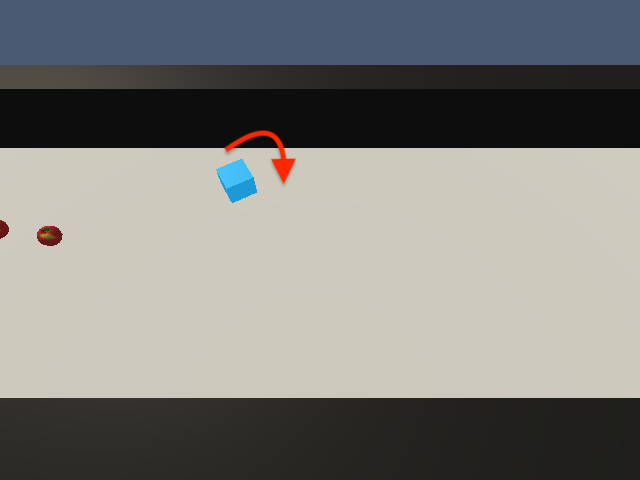}
		}
		\vspace{0.01cm}
		\subfloat[$t=48$\label{fig:result3_3}]{
			\includegraphics[clip,width=.49\linewidth]{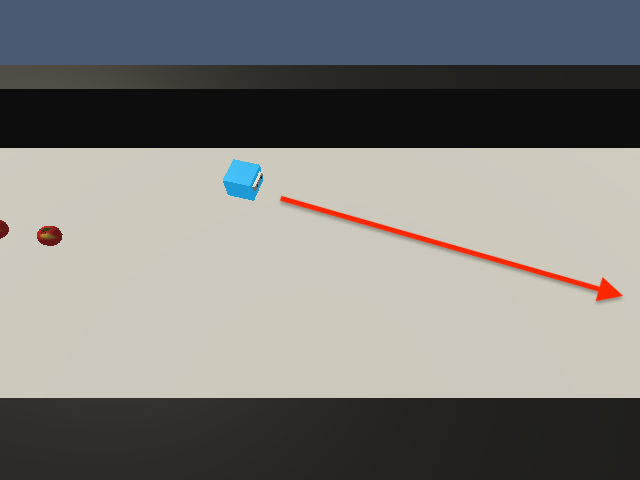}
		}
		\subfloat[probability \label{fig:result3_4}]{
			\includegraphics[clip,width=.49\linewidth]{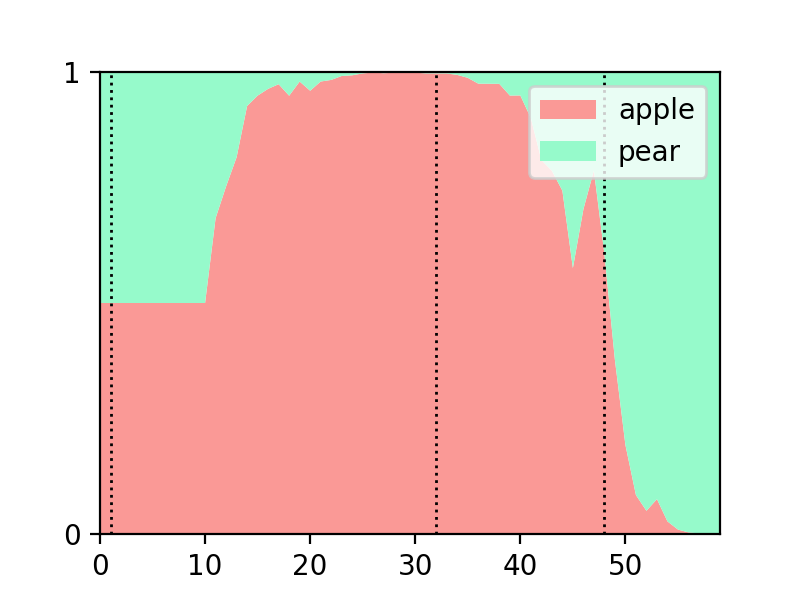}
		}
		\caption{
			The actor showed misleading behavior in episode 2.
			(a) The policy of the actor was trained to get a pear,
			but the actor first began moving toward the apples.
			(b) The actor changed its direction to the opposite side.
			(c) The actor moved to the right. The observer could not observe what the actor was moving toward. 
			(d) The PublicSelf model first inferred that it was more likely that the actor intended to get an apple when the actor approached the apple.
			Then, the probability decreased as the actor turned to the opposite side.
		}
		\label{fig:result2}
	\end{center}
\end{figure}

\begin{figure}[tbp]
	\begin{center}
		\subfloat[$t=1$\label{fig:result2_1}]{
			\includegraphics[clip,width=.49\linewidth]{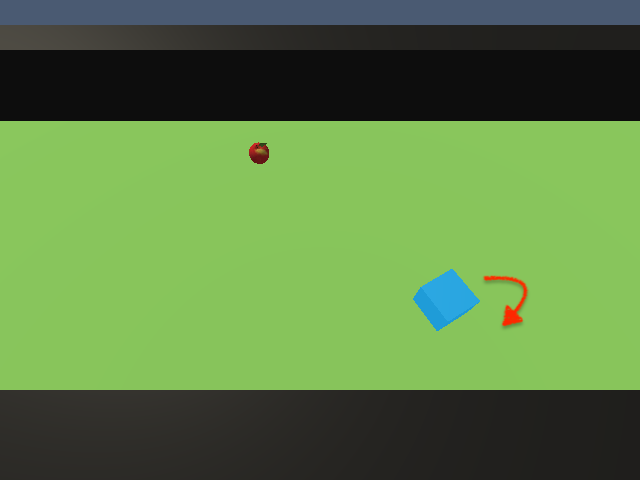}
		}
		\subfloat[$t=1$ (actor's perspective)\label{fig:result2_2}]{
			\includegraphics[clip,width=.49\linewidth]{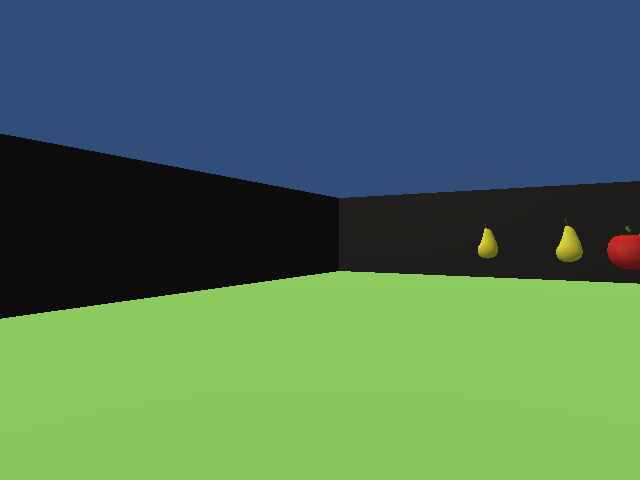}
		}
		\vspace{0.01cm}
		\subfloat[$t=10$\label{fig:result2_3}]{
			\includegraphics[clip,width=.49\linewidth]{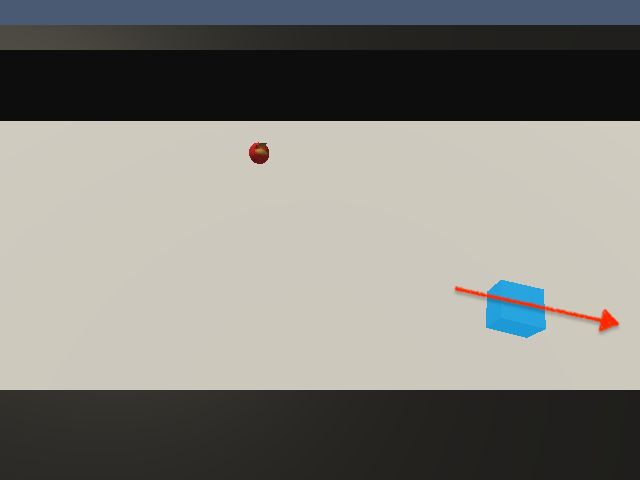}
		}
		\subfloat[probability\label{fig:result2_4}]{
			\includegraphics[clip,width=.49\linewidth]{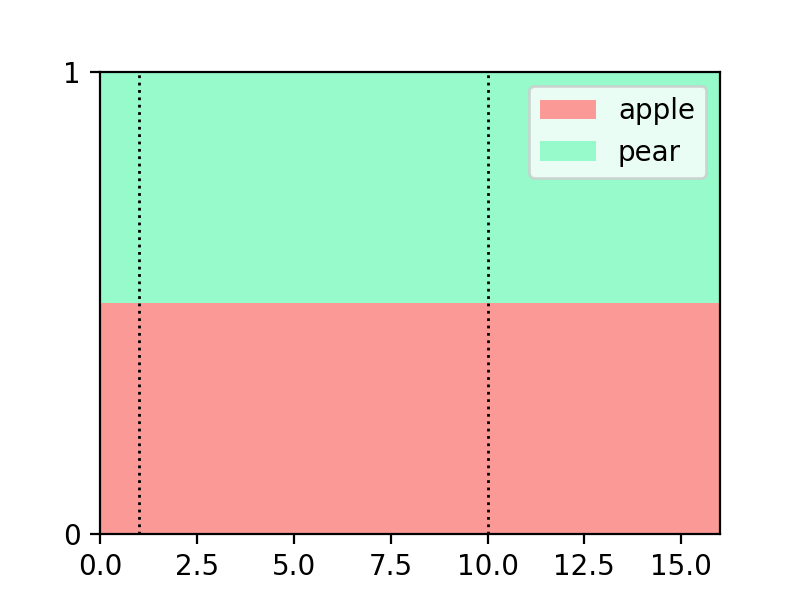}
		}
		\caption{
			Episode 3.
			In this episode, there was an apple at the middle of observer's perspective.
			In addition, there was an apple and two pears at the bottom right of the field,
			which the observer could not observe.
			(a) First, the actor faced the upper-right area and began turning clockwise.
			(b) An apple and two pears were in the sight of the actor at a time point $t=1$,
			but could not be seen from the observer's perspective.
			(c) The actor went toward the apple at the bottom right of the field, and disappeared
			from the observer's perspective.
			(d) The PublicSelf model observer could estimate the actor's intention because the observer did not
			know what the actor was moving toward.
		}
		\label{fig:result3}
	\end{center}
\end{figure}

Figures \ref{fig:result1}, \ref{fig:result2}, and \ref{fig:result3} show the conditions and estimated probabilities of 
the actor's intentions by the PublicSelf model in three episodes.
The actor's policy was $\pi_a$ in episodes 1 and 3, and $\pi_p$ in episode 2.

In episode 1,
the actor showed a behavior that made it relatively simple for an observer to infer the actor's intention.
The actor first faced to the bottom left, turned counterclockwise, and moved to an apple.
Suppose you were the observer inferring the actor's intention.
You could not judge the actor's intention while the actor is turning,
and would think the actor intended to get an apple after the actor began to move forward toward the right apple.
The probability of the intention inferred by the PublicSelf model rapidly increased when the actor began to move forward,
which seemed to meet our intuitive image.

In episode 2, 
the actor showed misleading behavior.
The actor moved toward the apples even though the actor's actual target was the pears.
Then, the actor changed its direction to the right and disappeared from the observer's sight.
The PublicSelf model first estimated that it was more likely that the actor intended to get apples,
and then greatly decreased the probability when the actor turned its back on the apples.

In episode 3, 
the observer obtained very poor information.
The actor actually observed two pears and an apple in their view, and moved toward the apple.
However, because the observer could not observe what the actor was moving toward, the observer could not infer the actor's intention.
The actor needed to consider the asymmetry of the information between the actor and the observer to estimate the observer's inference.
The PublicSelf model can take into account the fact that the observer does not know what exists in the right field.
Thus, we could obtain the result that the probabilities of $\iota_a$ and $\iota_p$ were even.

\newcommand{\argmax}{\mathop{\rm argmax}\limits}

\newif\ifja

\section{Comparison with people's judgment from observer's perspective}
\subsection{Experiment settings} 

In order to verify whether it is possible to use the PublicSelf model to actually comprehend people's utterances,
we conducted an experiment to compare the inferences of the PublicSelf model with the judgments of people 
who actually observed the agent's behavior from the observer's perspective.

\begin{figure}[tbp]
  \begin{center}
	\includegraphics[clip,width=0.65\linewidth]{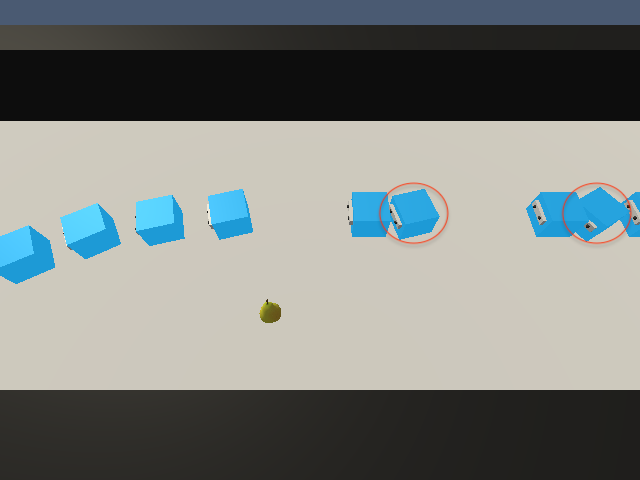}
	  \caption{Actor's behavior in episode 4. 
	   The actor appeared from the right and moved to the left, passing by a pear.
	  The actor twice moved toward the pear but soon changed direction, which could also mislead the mind-reading participants.}
  	\label{fig:ep4}
  \end{center}
\end{figure}

In this experiment, we prepared six episodes, including the ones shown in subsection \ref{3.1}.
The six episodes could be classified into three types: simple, blind, and misleading. 
The simple type included episode 1. 
In the simple episodes, the actor soon caught sight of its target and headed directly for it.
There were fewer factors for the participants to consider than in the episodes of the other types.
The blind type included episode 3. The actor seemed to head for something, but the observer could not observe the object toward which the actor was heading.
The misleading type included episodes 2 and 4 (Fig. \ref{fig:ep4}). 
There the actor showed misleading behavior by first heading for either apples or pears, but ignored them and moved in a different direction.
Misleading behavior often appears when the actor does not catch the actor's target in sight and searches for it.
Each type had two episodes.
The actor's policy was $\pi_a$ in one episode, and $\pi_p$ in the other.

\begin{figure}[tbp]
  \begin{center}
  	\includegraphics[clip,width=0.8\linewidth]{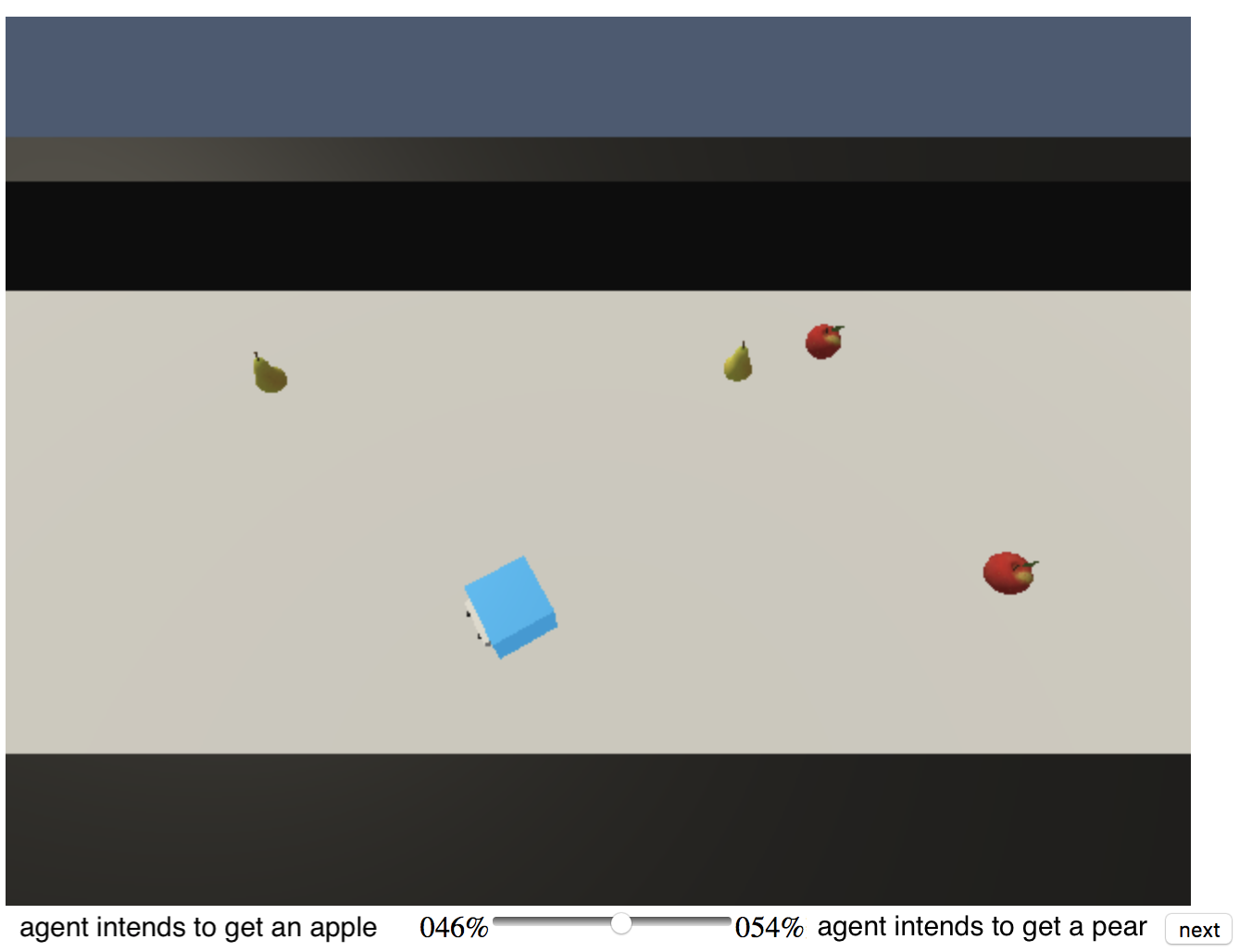}
	  \caption{User interface for participants. 
	  Participants indicated the probability by moving a slider.
	  The sentences were in Japanese in the actual experiment.}
  	\label{fig:ui}
  \end{center}
\end{figure}

Participants watched videos from the observer's view frame by frame,
and were asked to estimate the probability of the agent's intention for every frame on web browsers (Fig. \ref{fig:ui}).
The participants watched the eight episodes in a randomized order.
When the actor spawned in the observer's sight at the beginning of each episode, 
a red arrow indicated the direction of the actor.

The participants were 11 undergraduate and graduate students, eight male and three female.

Before the experiment, instructions were given to the participants.
An experimenter first showed the participants the whole area of the field, the observable areas of the observer and actor, 
and the appearance of the actor.
Then, the participants received an explanation of the use of the user interface,
and were given four points to note for the experiment as follows:
(i)~There were always one actor, two apples, and two pears on the field.
(ii)~The only possible intentions for the actor were either $\iota_a$ or $\iota_p$.
(iii)~The actor did not know where the fruits were on the field at the beginning of each episode.
(iv)~The initial positions of the actor and the fruits were randomly determined at the beginning of each episode independently of the actor's intention.
Instruction (iv) was given so that the participants considered the initial probabilities of the actor's intention to be even. 

In spite of instruction (iv), two female participants related the actor's spawn points to the actor's intention.
In other words, they thought that it was more likely that the actor intended to get an apple when the actor just spawned in front of an apple.
This suggests that a situation alone can have a strong influence on people's mind-reading even if the behavior of the actor had not been observed.
In order to consider the inference of an actor's intention from the actor's actions, however,
the two participants were excluded from the analysis in this paper.
Considering the influence of the situation is one direction for future work. 

\subsection{Results} 
\begin{figure}[tbp]
  \begin{center}
  	\includegraphics[clip,width=0.6\linewidth]{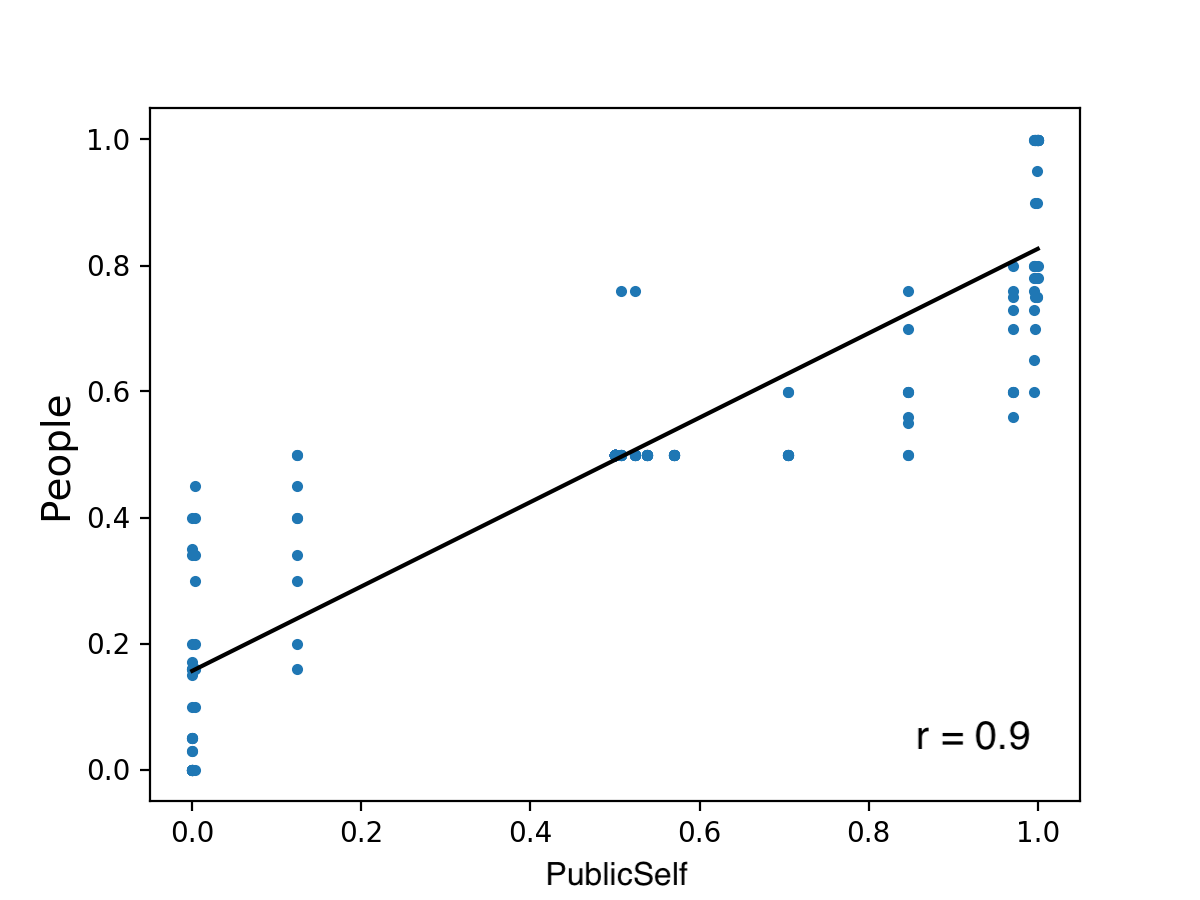}
	  \caption{Comparison of probabilities that actor has intention $\iota_a$
	  estimated by PublicSelf model and people in simple episodes.
	  Each dot represents an answer of a participant and the PublicSelf model in a step.
	  There was a strong and significant correlation between the people and the PublicSelf model.} 
  	\label{fig:typeA}
  \end{center}
\end{figure}

Figure \ref{fig:typeA} shows the relationship between the probabilities estimated by the PublicSelf model and participants in episodes of the simple type.
There was a strong and significant correlation ($r = 0.90$) between the estimation of the probability estimations.

\begin{figure}[tbp]
  \begin{center}
  	\includegraphics[clip,width=0.5\linewidth]{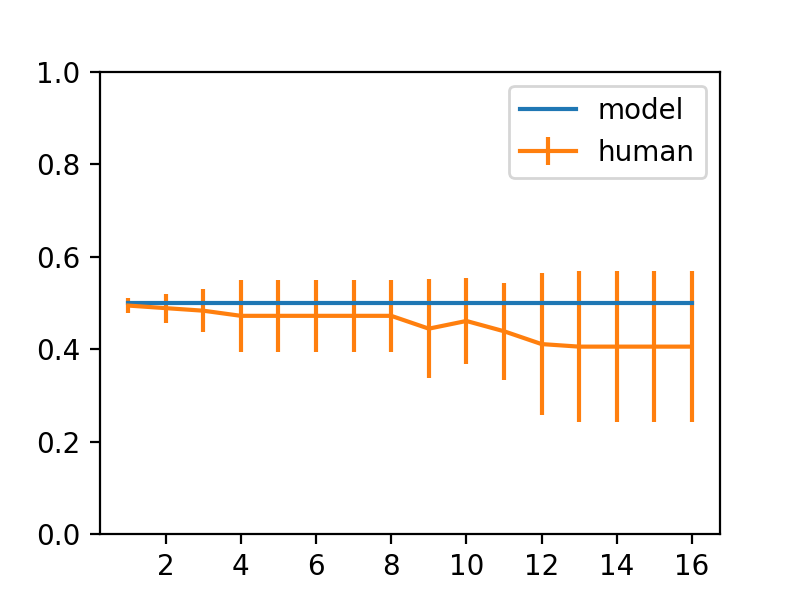}
	  \caption{Estimated probability that actor had intention $\iota_a$ for each step in episode 3.
		  Error bars show standard deviations of people's answers in a step. }
  	\label{fig:typeB}
  \end{center}
\end{figure}

In the blind episodes, the PublicSelf estimated that the probabilities were even at every step. 
However, in episode 2, three participants estimated that the actor had intention $\iota_p$ with a higher probability
after the actor began to move to the right of the field (Fig \ref{fig:typeB}).
One of the participants explained that
he estimated the probability of $\iota_p$ higher because 
the actor seemed to find its target on the right of the field, where a pear existed with a higher probability than an apple, 
because an apple already existed in the middle of the field.
This is a logical inference, but it was impossible for the actor to infer the participants' thought in the settings of this experiment 
because the actor had never observed the apple at the center of the field and did not know that the apple was there.
Information asymmetry between an actor and observer can sometimes cause this kind of insoluble problem.

\begin{figure}[tbp]
	\begin{center}
		\subfloat[Episode 2]{
			\includegraphics[clip,width=.5\linewidth]{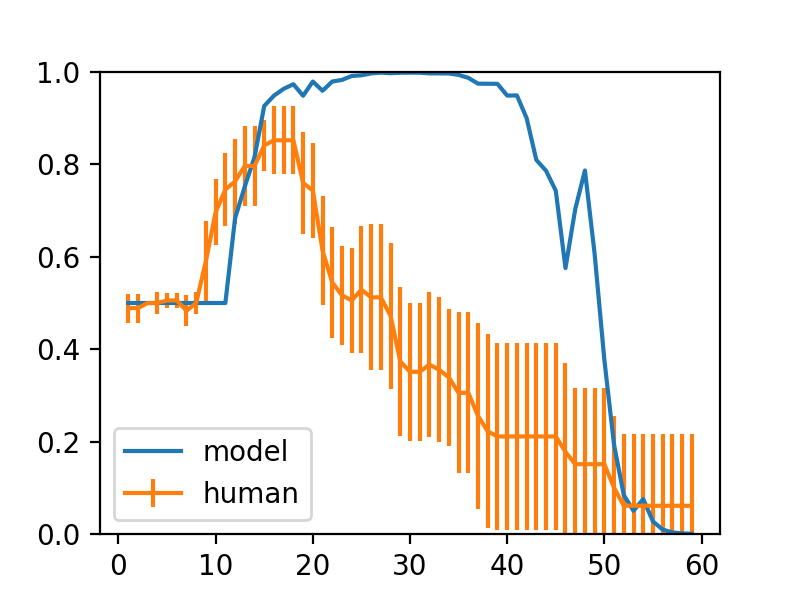}
		}
		\subfloat[Episode 4]{
			\includegraphics[clip,width=.5\linewidth]{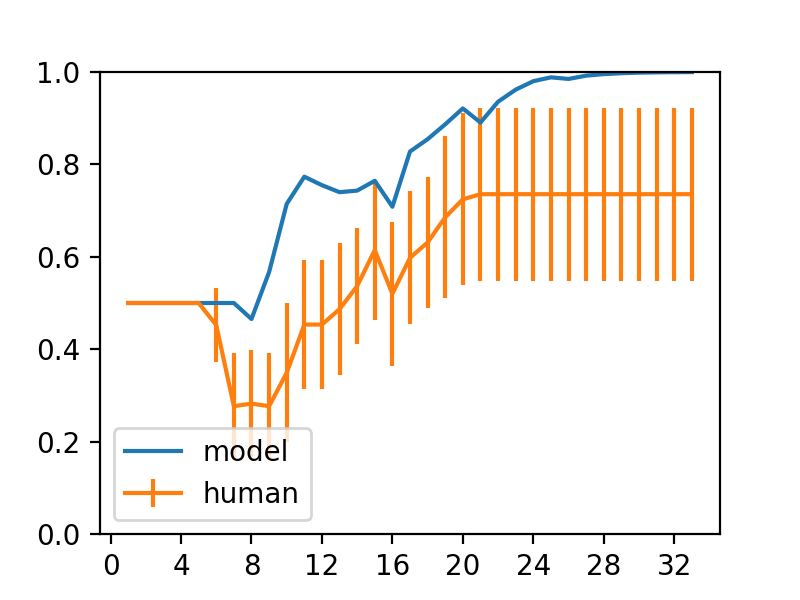}
		}
		\caption{
			There were gaps in the probabilities estimated by the people and PublicSelf model in the misleading episodes.
			In episode 2 (left), the probability of the people's estimation declined much more quickly than that of the PublicSelf model.
			In episode 4 (right), the probability of the people's estimation did not become 1.0 despite the fact that
			the actor seemingly ignored a pear, knowing that the pear was at the bottom of the field.}
		\label{fig:typeC}
	\end{center}
\end{figure}

Figure \ref{fig:typeC} shows the probabilities estimated by both people and the PublicSelf model in the misleading episodes. 
There was a significant but not strong correlation ($r = 0.46$).
There were two possible factors that led to the gaps between the estimations by the people and PublicSelf model.

One factor was that the people fed back the agent's behavior to the inference of the actor's beliefs.
When the actor changed direction, some participants let out a gasp of astonishment
and reported that it seemed the actor did not notice that the pear was there or was losing sight of the pear.
In our implementation, the update of the actor's belief was deterministic, 
but people fed back the actor's behavior to the inference of the actor's belief.
A probabilistic inference of the actor's belief could narrow the gap between the model and people.

Another possible factor is that the participants could have suspected the intentionality of the actor's behavior. 
In both of the misleading episodes, the actor much more frequently chose the action "do nothing" or 
alternatively selected the actions "turn clockwise" and "turn counterclockwise" compared to the simple episodes.
Because $\pi_a$ and $\pi_p$ were probabilistic, the actor sometimes, especially when it did not find its target, lost consistency and rationality of behavior.
In a questionnaire, some participants indicated that they felt anxious when the actor showed such inconsistent and irrational behaviors
and found it difficult to choose from the two alternatives $\iota_a$ and $\iota_p$.
The rationality of an agent is a necessary requirement for people to consider 
the agent to be an entity to which they can attribute mental states~\cite{rationality}.
The participants might have doubted the assumption that the actor had either intention $\iota_a$ or $\iota_p$ and 
answered with a probability of approximately fifty-fifty after losing confidence on their answers.
The confidence of people's judgments could also be related to whether they used context-dependent expressions for the actor.
Thus, it should be important to consider the strength of the intentionality that people feel from the actor's behavior,
as well as the probabilities of the candidates for the actor's intention.

\section{Conclusion}
We proposed the PublicSelf model, 
which is a model of an actor that infers the mental states that an observer watching the actor's behavior attributes to the actor. 
We implemented the PublicSelf model for an actor that learned its policy by deep reinforcement learning in a simulated environment.
Verification of the PublicSelf model showed that the model could infer the attributed intention of the agent 
in scenes where people could feel a certain intentionality from the actor's behavior.

Our simulation has limitations such as people's eyesight is fixed and predefined,
which can be problematic toward the application of the PublicSelf model to real-world scenario.
We are planning further investigation into the implementation of the PublicSelf model for a real-world robot.
Moreover, the experiment suggested that the RL agent's intentionality had an effect on people.
An investigation of people's cognitive process in understanding an RL agent's behavior  
is important future work for the realization of effective human RL-agent interaction.

\bibliographystyle{ACM-Reference-Format}
\bibliography{bib}

\end{document}